\documentclass[conference]{IEEEtran}
\IEEEoverridecommandlockouts
\usepackage{cite}
\usepackage{amsmath,amssymb,amsfonts}
\usepackage{graphicx}
\usepackage{textcomp}
\usepackage{xcolor}
\usepackage{csquotes}
\usepackage{dirtytalk}
\usepackage{tcolorbox}
\usepackage{float}
\usepackage{orcidlink}
\usepackage{multirow}%
\usepackage{mathrsfs}%
\usepackage{microtype}
\usepackage{booktabs}%
\usepackage{algorithm}%
\usepackage{algorithmicx}%
\usepackage{algpseudocode}% newer
\usepackage{array}
\usepackage{paralist}
\usepackage{tabularx} % For better table formatting, if needed
\usepackage{ragged2e} % For \RaggedRight in p-columns if using tabularx
\usepackage{tikz} % For drawing diagrams, though mermaid is text-based
\usepackage[inline]{enumitem}
\newcolumntype{Y}{>{\raggedright\arraybackslash}X} % ragged-right X

\def\BibTeX{{\rm B\kern-.05em{\sc i\kern-.025em b}\kern-.08em
    T\kern-.1667em\lower.7ex\hbox{E}\kern-.125emX}}
\begin{document}

\title{A Pattern Language for Resilient Visual Agents
% Architectural Patterns for Resilient Visual Agents*\\
% {\footnotesize \textsuperscript{*}Note: Sub-titles are not captured for https://ieeexplore.ieee.org  and
}
\author{\IEEEauthorblockN{Habtom Kahsay Gidey\orcidlink{0000-0001-5802-2606}, Alexander Lenz, and Alois Knoll\orcidlink{0000-0003-4840-076X}}
\IEEEauthorblockA{\textit{Technische Universität München, München, Germany} \\
\{habtom.gidey, alex.lenz, knoll\}@tum.de}
}
\maketitle
\begin{abstract}
Integrating multimodal foundation models into enterprise ecosystems presents a fundamental software architecture challenge. Architects must balance competing quality attributes: the high latency and non-determinism of vision language action (VLA) models versus the strict determinism and real-time performance required by enterprise control loops. In this study, we propose an architectural pattern language for visual agents that separates fast, deterministic reflexes from slow, probabilistic supervision. It consists of four architectural design patterns: (1) Hybrid Affordance Integration, (2) Adaptive Visual Anchoring, (3) Visual Hierarchy Synthesis, and (4) Semantic Scene Graph.
\end{abstract}

\begin{IEEEkeywords}
Amortized Inference, Architectural Design Patterns, Autonomous Agents, Cognitive Automation, GUI Automation, Robotic Process Automation, Vision Language Action Models, Visual Grounding
\end{IEEEkeywords}

\section{Introduction}
% --- Overview/context
Autonomous software agents are transitioning from rigid, API-based integrations to embodied systems that interact directly with Graphical User Interfaces (GUIs)~\cite{gideyUser2023,castrillo2025fundamentals}.
This shift is driven by cognitive automation, i.e., the need to automate multi-step, long-horizon tasks in legacy ERPs, expert systems, and air-gapped environments that demand high cognitive load and cause human error~\cite{clarke2018examination,herm2023impact,MacedoGRM24}.
In these environments, binary interfaces are frequently inaccessible, forcing agents to rely solely on visual perception to ground their decision-making and actions~\cite{leno2021robotic}.

% ----- -However ---
However, engineering such visually grounded agents presents an architectural dilemma characterized by a bimodal failure pattern~\cite{gidey2023towards,lecun2022path}.
On one hand, traditional robotic process automation (RPA) is computationally efficient but inherently brittle~\cite{leno2021robotic, yeh2009sikuli}.
Because these systems rely on fragile structural locators and static screen coordinates, even minor UI updates, such as layout shifts, resolution changes, or unexpected pop-ups, can disrupt execution and cause open-loop failures~\cite{leno2021robotic, yeh2009sikuli}.
On the other hand, end-to-end vision language action (VLA) models, such as UI-TARS and CogAgent, provide robust semantic understanding but introduce high latency, inference costs, non-determinism, and architectural entanglement~\cite{qin2025ui,hong2024cogagent,lu2024omniparser}.
This entanglement creates hidden technical debt within monolithic models; changes or fine-tuning risk catastrophic forgetting, the Changing Anything Changes Everything (CACE) anti-pattern, due to a lack of structural modularity, orthogonality, and isolation~\cite{sculley2015hidden,parnas1972criteria}.
Consequently, response times can lag several seconds per primitive action, e.g., a click, making monolithic VLAs unsuitable for enterprise real-time control loops. 
Ultimately, architects are forced to trade real-time control for semantic adaptability.

A viable design solution avoids this dichotomy through a hybrid architectural synthesis, reframing the semantic gap~\cite{lecun2022path} as a Sim2Real gap.
In this approach, agents plan within an idealized semantic world but execute actions in a noisy, partially observable digital reality.
Consequently, they must be designed not as rigid scripts, but as embodied agents that interact with GUI ecosystems via virtual percepts and effectors, demanding robust visual grounding and reasoning.
Drawing on principles from behavior-based robotics~\cite{brooks2003robust}, this requires a hierarchical architecture. 
By decoupling fast, deterministic grounding (System 1~\cite{kahneman2011thinking}) from the slow, probabilistic cognitive layer (System 2, encompassing the VLA supervisor for perceptual recovery and the semantic world model for long-horizon planning), this architecture bridges the gap between long-horizon cognitive planning and primitive sensorimotor execution.

% --- contribution ---
Rather than proposing a new perception model, this paper addresses a fundamental software architecture problem, tackling how to structure and compose heterogeneous components with conflicting quality attributes, namely, deterministic low-latency control and probabilistic high-latency semantic reasoning, into a dependable runtime architecture. 
The contribution to the software architecture community, therefore, lies in architectural decomposition, responsibility allocation across control loops, explicit component coordination, and the resolution of critical quality-attribute trade-offs, involving latency, reliability, explainability, and cost. 
We capture these recurring solutions in a pattern language for resilient visual agents, introducing four architectural design patterns: (1) Hybrid Affordance Integration, (2) Adaptive Visual Anchoring, (3) Visual Hierarchy Synthesis, and (4) Semantic Scene Graph. 
These patterns are synthesized into a reference architecture that isolates high-latency, probabilistic models within a supervisory layer. 
%while allowing low-latency execution components to operate efficiently without sacrificing semantic adaptability.
This hierarchical approach allows low-latency execution components to operate efficiently, ultimately balancing the semantic adaptability of end-to-end vision-based agents with the speed, auditability, and operational stability of traditional automation.

% ----- structure 
% The paper is structured as follows. Sect.~\ref{architecturalForces} sets the fundamentals and architectural forces. Sect.~\ref{Patterns} details the hierarchical pattern language. Sect.~\ref{evaluation} presents the evaluation, and Sect.~\ref{Discussion} discusses implications. Finally, Sect.~\ref{Conclusion} concludes the study with an outlook.
% % % % % % % % % % % % 
% % % % % % % % % % % % 
% % %  Section 2  % % % 
\section{Background and Related Work}\label{sec:background} 
%<background>
Autonomous agents operating on graphical user interfaces increasingly target screen-based enterprise environments lacking reliable APIs, DOM structures, or stable automation hooks, for example, legacy desktop applications, remote desktop sessions, or virtualized systems. 
In such settings, agents must visually perceive and act upon the interface much like human users~\cite{gidey2023towards}. 
This creates a demanding engineering context where perceptual uncertainty, UI variation, latency constraints, and operational reliability are first-class concerns.

Historically, GUI automation has been dominated by deterministic approaches such as scripts, macros, and RPA. 
While efficient, auditable, and suited for repetitive workflows, they are often brittle under UI drift, layout shifts, or missing selectors. 
More recently, VLA models, such as CogAgent and UI-TARS, have demonstrated the ability to interpret raw screen pixels and generate semantically informed actions~\cite{qin2025ui,hong2024cogagent,castrillo2025fundamentals}. 
However, while improving adaptability, VLAs introduce new challenges regarding high latency, probabilistic behavior, cost, and reduced predictability in production.

This tension and sharp increase in complexity reveal a fundamental software architecture problem, not merely a model-selection one.
Architecture provides established mechanisms, e.g., decomposition, explicit coordination, layered control, and reusable patterns, to structure component responsibilities and interaction boundaries~\cite{parnas1972criteria}. 
By systematically constraining the design space to balance conflicting forces, architectural patterns provide behavioral guarantees~\cite{DBLP:conf/facs2/GideyCM19,marmsoler2019interactive,gidey2018factum}. 
They translate architectural assumptions and design goals into explicit structural boundaries, i.e., components and connectors.  

%<related work>
Although recent research actively explores agentic visual perception, it predominantly treats agents as monolithic, end-to-end models~\cite{lu2024omniparser}. 
Existing literature lacks architectural approaches that combine heterogeneous models and deterministic execution components under real-time constraints. 
To address this gap, our work synthesizes established interdisciplinary paradigms, i.e., the ecological approach to visual perception~\cite{gibson1979ecological}, Kahneman's~\enquote{System 1 and System 2} dual-process theory~\cite{kahneman2011thinking}, the layered subsumption architecture from robotics~\cite{brooks2003robust}, and the autonomic MAPE-K control loop~\cite{kephart2003vision}. 
By applying these foundations, the subsequent sections propose a hierarchical pattern language and reference architecture for resilient user-like visual agents~\cite{gideyUser2023}.
% % % % % % % % % % % % 
% % % % % % % % % % % % 
% % %  Section 3  % % % 
%\section{Fundamentals and Architectural Forces}\label{architecturalForces}
\section{Architectural Forces}\label{architecturalForces}
%\section{Architectural Forces and Reference Architecture}\label{architecturalForces}
To bridge the Sim2Real gap inherent in visual agents, we must move beyond viewing the agent as a script and instead model it as an embodied agent.
This perspective shifts the architectural requirements from simple~\enquote{\textit{execution}} to~\enquote{\textit{perception, control, and correction.}}
In this section, we outline the foundational assumptions and the competing architectural forces that drive the identification and formulation of the architectural pattern language.

\subsection{The Embodied Agent Paradigm} 
Traditional automation assumes an~\enquote{\textit{open-loop}} model, i.e., the agent sends a command, such as~\texttt{click(x,y)}, and assumes the state changes. However, in non-instrumented environments, such as VNC or Citrix-based virtualized desktops, this assumption is fallacious due to network latency, pop-ups, or rendering lag.
We adopt the MAPE-K,~\textit{Monitor-Analyze-Plan-Execute-Knowledge} loop, from autonomic computing~\cite{kephart2003vision} to model the agent as an adaptive self-healing system~\cite{gidey2023modeling}, and integrate it with~\textit{subsumption architecture} principles from robotics~\cite{brooks2003robust}. In this model, the agent possesses:
\begin{itemize}
    \item \textbf{Visual Proprioception:} The ability to verify its own actions via optical flow, for instance,~\enquote{\textit{Did the button depress?}}.
    \item \textbf{Active Perception:} The autonomy to perform epistemic actions, such as scrolling, to resolve uncertainty~\cite{gibson1979ecological}.
    \item \textbf{Hierarchical Control:} A fast reflex layer for execution, System 1, and a slow deliberative cognitive layer for planning and recovery, System 2.
\end{itemize}

\subsection{Architectural Forces}
The design of any autonomous agent in legacy enterprise environments, such as ERP ecosystems, involves resolving four fundamental, often conflicting, forces: 

\begin{itemize} [leftmargin=*, noitemsep, topsep=3pt]
    \item \textbf{Time:} This force relates to the latency vs. semantics trade-off. High-level reasoning requires massive VLA models, which incur high latency, usually 2 to 5 seconds for inference, extending to $\sim$10s with network round-trips~\cite{baechler2024screenai}. However, precise interaction, for example, clicking a moving button, requires millisecond-level feedback loops. An architecture that relies solely on VLAs (end-to-end) will act too slowly for real-time control, while one that relies solely on heuristics (RPA) lacks semantic understanding. \\
    \textit{Driver:} The architecture must decouple grounding, fast, from reasoning/planning, slow.
    
    \item \textbf{Cost:} This relates to the economic trade-off between operational expenditure, i.e., inference cost, and cognitive capability. Executing high-parameter VLA models for every atomic primitive action, for example, moving the cursor or verifying a state change, incurs prohibitive compute overhead and API fees. While traditional RPA scripts are computationally trivial, relying on inexpensive local CPU cycles, an end-to-end VLA approach scales cost linearly with every interaction. In high-volume enterprise workflows, such as batch invoice processing, paying for deep semantic reasoning at every step is financially unviable. Conversely, traditional RPA is cheap to execute but expensive to maintain when UI changes break the automation. \\ 
    \textit{Driver:} The architecture must amortize the cost of intelligence by translating expensive semantic decisions into cheap, reusable local reflexes, invoking the costly probabilistic supervisor only upon perceptual failure or UI drift.
    
    \item \textbf{Reliability:} This represents the determinism vs. adaptability paradox. Enterprise environments demand highly repeatable and reliable workflows, i.e., \textit{determinism}. An agent must not hallucinate a~\enquote{\textit{Delete}} button where none exists. Conversely, these environments are dynamic. Purely deterministic approaches fail on updates, whereas purely probabilistic approaches fail in terms of reliability. \\
    \textit{Driver:} The architecture must provide a deterministic reliability layer that can be dynamically updated by a probabilistic supervisor, i.e., System 2 repairing System 1 via self-healing mechanisms~\cite{gidey2023modeling}.
    
    \item \textbf{Explainability:} When a~\textit{black box} model fails, diagnosing whether the failure was perceptual,~\textit{didn't see the button}, or logical,~\textit{chose the wrong button}, is often impossible. This lack of explainability and failure isolation is unacceptable in industrial settings. \\
    \textit{Driver:} The architecture must produce an explicit, queryable world model,~\textit{scene graph}, to ensure that decision-making logic is transparent and observable.
\end{itemize}
% % % % % % % % % % % % 
% % % % % % % % % % % % 
% % %  Section 3  % % % 
\section{A Hierarchical Pattern Language}\label{Patterns}

\begin{figure}[!htbp]
\centerline{\includegraphics[scale=0.51]{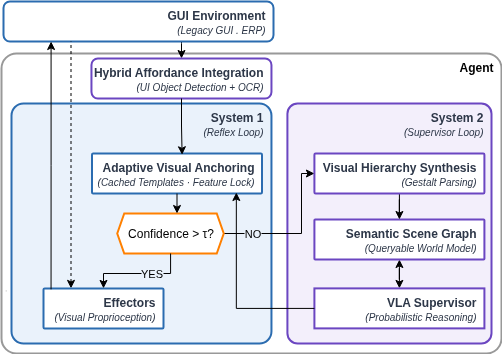}}
\caption{The hierarchical reference architecture showing separation of reflex (UI Objects) and reasoning (VLA / System 2).}
\label{fig:pattern1}
\end{figure}

To resolve these forces, we propose a reference architecture, Fig.~\ref{fig:pattern1}, composed of three asynchronous control loops, i.e.,~\textit{Reflex, Structural, and Supervisor}, implemented via four architectural patterns. This structure enables amortized inference, in which high-cost reasoning is invoked only to repair low-cost reflexes. 
% \begin{figure*}[!htbp]
% \centerline{\includegraphics[scale=0.58]{img/raPattern.png}}
% \caption{The hierarchical reference architecture showing separation of reflex (UI Objects) and reasoning (VLM / System 2).}
% \label{fig:pattern1}
% \end{figure*}

This section details the four architectural patterns that instantiate the reference architecture. Each pattern defines a reusable solution at the subsystem and interaction level by specifying structural responsibilities, coordination boundaries, and communication rules among components~\cite{gamma1994design,DBLP:conf/facs2/GideyCM19}. The contribution therefore lies not in prescribing specific perception models or algorithms, but in structuring how low-latency reflexes, structural mediation, and high-latency supervisory reasoning are composed to resolve competing quality attributes. Due to space limitations, we primarily focus on how these architectural patterns address conflicting forces and guide the associated design rationale and architectural design decisions. The pattern specification adopts the canonical pattern documentation format~\cite{gamma1994design}, i.e.,~\textit{Context, Problem, Solution}, with explicit attention to the architectural forces each pattern addresses.

\subsection{Hybrid Affordance Integration Pattern}
\subsubsection*{Context} The agent operates in a~\textit{noisy} visual environment where distinct sensory modalities often conflict. As an example, an object detection model might classify a colored rectangle as a~\enquote{\textit{button,}} while an OCR model reads non-actionable text, for instance,~\enquote{\textit{Version 2.0}} text, inside the rectangle.
\subsubsection*{Problem} Relying on a single modality, vision or text, creates a single point of failure, and a~\enquote{\textit{bag of detections}} approach fails to resolve sensory conflicts, both leading to the~\enquote{\textit{hallucination of affordance}}, where the agent attempts to interact with non-functional UI elements~\cite{gidey2025affordance}, i.e.,~\textit{sensory hallucination.}
\subsubsection*{Solution} Implement a multimodal perception-fusion layer acting as a perceptual arbiter component. This architectural pattern dictates the parallel execution of independent sensor components, such as a UI object detector and OCR, into a parallel processing structure and defines how their outputs are fused into a normalized, uncertainty-aware affordance interface~\cite{gidey2025affordance}. The resulting subsystem mediates between raw perceptual signals and downstream control loops, enabling more robust and semantically grounded interaction with the UI.
%\begin{itemize*} [label={}]
\begin{itemize}[leftmargin=*, noitemsep, topsep=3pt]
\item \textit{Parallel Perception:} Run lightweight local models in parallel streams, a fast UI objects detector for widget localization, and an OCR engine for semantic text extraction.
\item \textit{Geometric Fusion:} Project both outputs onto a shared coordinate plane and align the bounding boxes.
\item \textit{Conflict Resolution:} If vision and text disagree, the arbiter downgrades the confidence score, flagging the object as~\enquote{\textit{uncertain}} to trigger active verification, for instance, hovering, rather than execution. By filtering out noise at the source, this pattern acts as the data cleaning filter for the architecture, ensuring that the downstream world model is populated only with multi-verified entities, directly addressing the explainability force.
\end{itemize}

\subsection{Adaptive Visual Anchoring Pattern}
\subsubsection*{Context} Enterprise execution requires millisecond-level reaction times for primitive action control, i.e., visual proprioception, and strict determinism for reliability and safety. End-to-end vision models are too slow for this loop, while static templates are too brittle to handle UI updates.
\subsubsection*{Problem} The agent needs the speed of a hash-map lookup, finding an element instantly, but also the flexibility of a foundation model, handling icon updates. Relying solely on templates leads to brittle failure; relying solely on VLAs leads to unacceptable latency, i.e.,~\textit{the drift-latency dilemma}.
\subsubsection*{Solution} Implement a fallback-routing connector that establishes a structural boundary between a fast, deterministic Local Cache Component (System 1) and a deliberative VLA Supervisor Component (System 2). Architecturally, this pattern separates runtime grounding from supervisory repair and defines the hand-off contract between the Reflex loop and the Supervisor loop. The connector governs when execution remains within the low-latency path and when control is escalated to supervisory reasoning, thereby preserving responsiveness while enabling recovery from uncertainty, drift, and failure.
%\begin{itemize*}[label={}]
\begin{itemize}[leftmargin=*, noitemsep, topsep=3pt]
\item \textit{System 1 - Runtime:} The agent grounds interaction using cached visual anchors, templates, or feature embeddings. This occurs in $< 50$ms, driving the fast~\enquote{\textit{reflex loop.}}
\item \textit{The Drift Exception:} If the deterministic match confidence falls below a safety threshold, $\tau < 0.9$, the system halts execution rather than guessing.
\item \textit{System 2 - Supervisor:} The exception triggers the deliberative VLA supervisor, System 2. The VLA semantically locates the entity, for instance,~\enquote{\textit{Find the Save icon, even if it is green}}, crops the new visual template, and updates the System 1 cache. This resolves the latency-adaptability trade-off, allowing the agent to run at the speed of a script while retaining the adaptability of an LLM, effectively amortizing the high cost of intelligence over thousands of cache-hit interactions.
\end{itemize}

\subsection{Visual Hierarchy Synthesis Pattern}
\subsubsection*{Context} The output of sensory processing is a flat list of GUI elements. Without a structured UI representation, the agent suffers from relational ambiguity, unable to distinguish between identical elements based on context, for example, distinguishing the~\enquote{\textit{Edit}} icon in row A vs. row B of a datagrid.
\subsubsection*{Problem} The planner requires relative addressing, i.e., context, whereas the sensors provide absolute addressing, i.e., coordinates. A global layout shift moves a button to $(x+50, y)$, breaking absolute references and causing an ambiguity of reference problem.
\subsubsection*{Solution} % Reverse-engineer a structured DOM-like interface tree from the flat pixel stream using Gestalt Principles. Architecturally, this pattern inserts a structural mediation layer or structural adapter component that transforms flat percepts into a hierarchical UI representation consumable by planners.
Introduce a structural mediation layer that aggregates flat perceptual outputs into a hierarchical UI representation using Gestalt-inspired grouping principles. Architecturally, this pattern acts as a structural adapter component between low-level perception and downstream planning, transforming unstructured percepts into a planner-consumable interface hierarchy.
%\begin{itemize*}[label={}]
\begin{itemize}[leftmargin=*, noitemsep, topsep=3pt]
\item \textit{Layout Parsing:} The algorithm analyzes the spatial layout of detected affordances. It applies perceptual grouping rules, specifically~\textit{containment} and \textit{alignment}, to infer a tree structure.
\item \textit{Relative Addressing:} The system synthesizes logical groups,~\textit{forms, tables, and modals}, allowing the planner to issue robust commands like \texttt{click("Submit", "Login\_Form")}.
This mechanism provides~\textit{robustness to translation}. Even if the~\enquote{\textit{Login Form}} moves 500 pixels to the right, the internal parent-child structure remains invariant, ensuring the agent’s plan succeeds without modification.
\end{itemize}
\subsection{Semantic Scene Graph Pattern}
\subsubsection*{Context} Complex tasks require long-horizon planning. To avoid unsafe actions, for instance, deleting a file, the agent needs to understand the structural context before committing to it.
\subsubsection*{Problem} A flat list of elements captures the current state but lacks semantic and spatial relationships, preventing the agent from understanding context.
\subsubsection*{Solution} Construct a static, queryable Semantic Scene Graph. Architecturally, this pattern provides an explicit world-model subsystem that decouples decision logic from raw perceptual outputs and supports verification and observability.
%\begin{itemize*}[label={}]
\begin{itemize}[leftmargin=*, noitemsep, topsep=3pt]
\item \textit{The Graph:} Nodes represent semantic entities resulting from the third pattern; edges represent functional relationships and spatial constraints. 
\item \textit{Verification:} Before executing a high-risk plan, the agent traverses this queryable static graph to verify preconditions and identify safety violations explicitly. This transforms the scene graph from a passive semantic map into an explicit reasoning structure, addressing the reliability and explainability forces.
\end{itemize}
\section{Evaluation}\label{evaluation}
This study is validated using the scenario-based architecture analysis method (SAAM)~\cite{kazman1994saam}. Due to space constraints, this evaluation focuses specifically on resolving the latency and safety forces during a UI stress event, tracing the Adaptive Visual Anchoring pattern, System 1/2 handoff.

\subsection{Methodological Setup}
We define a simplified~\enquote{\textit{canonical legacy scenario}} derived from recurring maintenance challenges observed in non-instrumented financial ERPs, such as DATEV, and proprietary database front-ends, such as MS Access. 
The workflow tasks an autonomous agent with a high-frequency batch process:~\enquote{\textit{invoice approval,}} requiring it to validate a data field, a common task in document-based knowledge discovery~\cite{gidey2022document}, and click a specific~\enquote{\textit{Submit}} button. The stress event is a minor vendor update that introduces a UI perturbation (drift): the~\enquote{\textit{Submit}} button moves 50 pixels to the right, changes style (e.g., from blue to green), and a destructive~\enquote{\textit{Delete}} or~\enquote{\textit{Cancel}} button is introduced at the \emph{exact previous coordinates} of the~\enquote{\textit{Submit}} button. The quality goals are strict \textit{safety} (never click~\enquote{\textit{Delete}}) and \textit{latency} suitable for batch processing, $<1$s per record.

\subsection{Architectural Walkthrough}
We trace the execution of the workflow through the three architectural paradigms.

\textbf{Baseline A - Classic RPA:} The script operates open-loop, executing the hardcoded command~\texttt{click(x, y)}. Because the~\enquote{\textit{Delete}} trap now occupies the target coordinates, the agent unintentionally destroys the record, causing a catastrophic failure. The system lacks visual proprioception and assumes a static environment, violating the safety requirement.

\textbf{Baseline B - End-to-end VLA:} The system captures a screenshot, uploads it to the VLA, and prompts, e.g.,~\enquote{\textit{Click Submit.}} The model successfully identifies the semantic concept~\enquote{\textit{Submit}}, the new green button. However, the inference round-trip takes $\sim$10 seconds. For a batch of 1,000 invoices, this adds nearly 3 hours of processing overhead compared to RPA. While resilient, the system violates the latency and cost forces, making it economically unviable for high-frequency control loops.

\textbf{Proposed Architecture:} Our proposed architecture handles the scenario through amortized inference. First, the adaptive visual anchor pattern attempts to match the cached blue~\enquote{\textit{Submit}} template at the original coordinates. The match confidence drops below the safety threshold, $\tau < 0.9$, due to the color change and pixel shift. The reflex loop inhibits the click immediately, preserving safety. This reflex failure triggers the VLA Supervisor, System 2, which visually analyzes the new UI state, semantically locates the new green button, and updates the first layer's visual template cache and coordinates. The agent resumes execution, and for the remaining 999 invoices in the batch, executes at millisecond speeds using the updated reflex loop.

\subsection{Trade-off Analysis}
The economic viability of the architecture follows from its asymmetric use of computation: low-cost reflexive execution handles the common case, while high-cost supervisory reasoning is invoked only under uncertainty, drift, or failure. This behavior can be expressed through an amortized cost and latency model:
% \begin{equation}
%     Cost_{avg} = Cost_{Reflex} + \left( P_{Drift} \times Cost_{Supervisor} \right) 
% \end{equation}
\begin{equation}
    \mathrm{Cost}_{\mathrm{avg}} = \mathrm{Cost}_{\mathrm{Reflex}} + \left( P_{\mathrm{Drift}} \times \mathrm{Cost}_{\mathrm{Supervisor}} \right) 
\end{equation}
% where $P_{Drift}$ is the probability of a UI perturbation or layout shift (e.g., $1\%$). Since the reflex loop ($Cost_{Reflex}$) incurs negligible computational cost and latency ($<50$\,ms), even if the VLA supervisor ($Cost_{Supervisor}$) takes $\sim$10 seconds and incurs API fees, a low $P_{Drift}$ ensures the average operational overhead approaches zero.

where $P_{\mathrm{Drift}}$ is the probability of a UI perturbation or layout shift (e.g., $1\%$). Since the reflex loop ($\mathrm{Cost}_{\mathrm{Reflex}}$) incurs negligible computational cost and latency ($<50$\,ms), even if the VLA supervisor ($\mathrm{Cost}_{\mathrm{Supervisor}}$) takes $\sim$10 seconds and incurs API fees, a low $P_{\mathrm{Drift}}$ ensures the average operational overhead approaches zero.
% where $P_{Drift}$ is the probability of a UI perturbation or layout shift. Since the reflex loop ($Cost_{Reflex}$) incurs negligible computational cost and latency (typically below 50\,ms), and the supervisor ($Cost_{Supervisor}$) is activated only rarely, the additional overhead introduced by supervisory reasoning remains low in expectation, even when a single supervisory intervention is comparatively expensive in both time and API cost. 
This makes the architecture economically viable for high-frequency batch processing while maintaining high resilience.
%
% % % % % % % % % % % % 
% % % % % % % % % % % % 
% % %  Section 56 % % % 
% \section{Discussion}\label{Discussion}
% Transitioning from traditional RPA to embodied visual agents introduces a testing paradox, i.e. traditional output-assertion testing fails when an agent autonomously self-heals using probabilistic models. Our architecture mitigates this paradox by explicitly separating the reflex and reasoning layers. Because the VLA updates System 1's deterministic cache rather than executing primitive actions directly, auditors can isolate the exact visual anchors the agent relies upon at runtime. The resulting state-transition graph serves as a queryable audit trail, transforming opaque model decisions into observable state changes. Consequently, organizations can shift from brittle pixel-level assertions to semantic contract testing, verifying scene graph boundaries without restricting the agent's probabilistic recovery mechanisms.
% % % % % % % % % % % % 
% % % % % % % % % % % % 
% % %  Section 7 % % % 
\section{Conclusion}\label{Conclusion}
This study addresses a recurring software architecture challenge in agent engineering: the dichotomy between brittle deterministic pre-scripted workflows and the semantic flexibility of slower vision models. 
To resolve this, we presented an architectural approach that constrains high-cost semantic reasoning to a~\enquote{\textit{supervisor loop,}} which repairs and guides a low-latency~\enquote{\textit{reflex loop.}}

By formalizing this approach into a pattern language for resilient visual agents, we transform the agent from an open-loop script into a closed-loop control system capable of active perception and autonomous recovery. 
Our feasibility analysis indicates that this architectural pattern preserves the speed and auditability of industrial automation while improving resilience through selective supervisory reasoning. 
Furthermore, by generating an explicit, queryable semantic scene graph, the architecture resolves part of the testing paradox of probabilistic models, enabling organizations to shift toward more robust semantic contract testing.
%Furthermore, by maintaining an explicit, queryable semantic scene graph, the architecture makes probabilistic perception more testable, enabling organizations to move toward robust semantic contract testing.

As an outlook, we plan to fully validate this architecture through quantitative evaluation across broader industrial case studies. 
Specifically, we will benchmark the architecture against state-of-the-art baselines, including pure VLA systems, to measure the statistical significance of latency reductions, resilience to UI drift, and overall operational robustness.

%\section*{Acknowledgment}
%The preferred spelling of the word ``acknowledgment'' in America is without an ``e'' after the ``g''. Avoid the stilted expression ``one of us (R. B. G.) thanks $\ldots$''. Instead, try ``R. B. G. thanks$\ldots$''. Put sponsor acknowledgments in the unnumbered footnote on the first page.

\bibliographystyle{IEEEtran}
\bibliography{bib/IEEEabrv,bib/main}

% IEEE conference templates contain guidance text for composing and formatting conference papers. Please ensure that all template text is removed from your conference paper prior to submission to the conference. Failure to remove the template text from your paper may result in your paper not being published.

\end{document}